\documentclass[a4paper,conference]{IEEEtran}
\usepackage{graphicx}
\usepackage{color}
\ifCLASSOPTIONcompsoc
    \usepackage[caption=false, font=normalsize, labelfont=sf, textfont=sf]{subfig}
\else
\usepackage[caption=false, font=footnotesize]{subfig}
\fi
\ifCLASSINFOpdf
\else
\fi
\usepackage{caption}

\usepackage{mathtools}
\newcommand{\ignore}[1]{}

\begin{document}
\title{Relative Attribute Classification \\ with Deep Rank SVM}


\author{\IEEEauthorblockN{Sara Atito Ali Ahmed and Berrin Yanikoglu}
\IEEEauthorblockA{Faculty of Engineering and Natural Sciences\\
Sabanci University\\
Istanbul, Turkey 34956\\
\{saraatito,berrin\}@sabanciuniv.edu
}
}

\maketitle

\begin{abstract}
Relative attributes indicate the strength of a particular attribute between image pairs. We introduce a deep Siamese network with rank SVM loss function, called Deep Rank SVM (DRSVM), in order to decide which one of a pair of images has a stronger presence of a specific attribute.  
%trained on relative based objective. 
%In this architecture, given a pair of images, we learn whether or not the primary image in the pair has a stronger presence of a specific attribute than the second image. 
%is optimized via unified end-to-end deep learning scheme. 
%BY
The network is trained in an end-to-end fashion to jointly learn the visual features and the ranking function.
We demonstrate the effectiveness of our approach against the state-of-the-art methods on four image benchmark datasets: LFW-10, PubFig, UTZap50K-lexi and UTZap50K-2 datasets.
DRSVM surpasses state-of-art in terms of the average accuracy across  attributes, on three of the four image benchmark datasets. 
%, with improvements reaching 1.21\% points
\end{abstract}

\IEEEpeerreviewmaketitle

\section{Introduction}
Identification and retrieval of images and videos with certain visual attributes are of interest 
%Visual attribute learning has been explored 
in many real-world applications, such as image search/retrieval \cite{kovashka2015whittlesearch,kovashka2017attributes}, video retrieval \cite{chen2014instructive}, image/video captioning \cite{pan2017video,yao2017boosting}, face verification \cite{kumar2009attribute}, and zero-shot learning \cite{fu2018recent,bansal2018zero}. 
Visual attribute learning is studied in particular for {binary} attributes that indicate the presence or absence of a certain semantic attribute (smiley, wearing eye glasses, etc.) \cite{ahmed2019within,zhuang2018multi}.

\begin{figure} [ht]
  \centering
  \subfloat{\includegraphics[width=\linewidth]{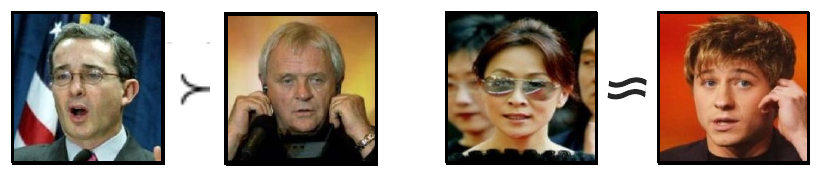}}
  \caption*{(a)}
  \hfill
 % \subfloat[(a)]{\includegraphics[width=\linewidth]{Figs/Test_LFW10.pdf}}
%  \hfill  
  \subfloat{\includegraphics[width=\linewidth]{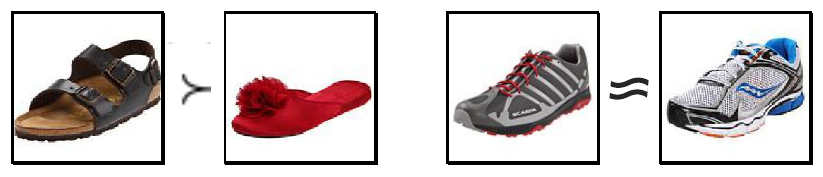}}
  \caption*{(b)}
%  \hfill
%  \subfloat[(b)]{\includegraphics[width=\linewidth]{Figs/Test_Zappos.pdf}}
  \caption{Samples of visual relative attributes from the training dataset. (a) Shows random samples from LFW10 dataset of \textit{mouth-open} attribute, and (b) Shows random samples from UTZap50K-2 dataset of \textit{sporty} attribute.}
  \label{fig:Compa} 
\end{figure}

Parikh and Grauman \cite{parikh2011relative} introduced \textit{relative attributes} with a formulation similar to that of Support Vector Machines (SVMs). 
The goal of relative attribute learning is to learn a function
which predicts the relative strengths of a pair of images regarding a given attribute (e.g. which picture is more smiling?). The network should be able to answer the
comparisons, with more/less/equal of the presence
of a specific attribute. 
Figure \ref{fig:Compa} shows the comparison for two separate data sets, for the attributes \textit{mouth-open} and \textit{sporty} from the LFW10 and UTZap50K-2 datasets. 

After the introduction of the problem, subsequent research approached the problem using either traditional features or deep learning approaches, as discussed in Section \ref{sec:relatedWorks}.  

In this paper, we present a deep learning system that can compare the given two images in terms of their strength regarding a particular attribute. Specifically, we propose a convolutional Siamese network 
%propose a novel deep relative attribute learning model 
using rank SVM loss function for the relative attribute problem. The main contributions of our proposed model are summarized as follows:
\begin{itemize}
    \item Proposing an end-to-end deep learning framework in which the network jointly learns the visual features and the rank SVM function, for relative attribute classification. 
    \item Demonstrating the effectiveness of the proposed framework by comparing to our baseline \cite{souri2016deep}, with  improvements of  6\%, 3\%, 2.65\%, and 1\% in average ranking accuracy for LFW-10, PubFig, UTZap50K-lexi, and UTZap50K-2 datasets, respectively.
    \item Surpassing the state-of-the-art results in LFW-10, PubFig, and UTZap50K-lexi datasets by about 2\%, 0.2\%, and 0.87\% and obtaining slightly lower results in UTZap50K-2 dataset.
\end{itemize}

The rest of this paper is organized as follows. A review of literature is presented in Section \ref{sec:relatedWorks}. Section \ref{sec:DRSVM} introduces our proposed method together using the deep rank SVM objective function. Description of the employed datasets, implementation details, along with extensive experimental results are discussed in Section \ref{sec:ExpRes}. Finally, the paper concludes in Section \ref{sec:conc}.

\begin{figure*}[ht]
    \centering
    \includegraphics[width=\linewidth]{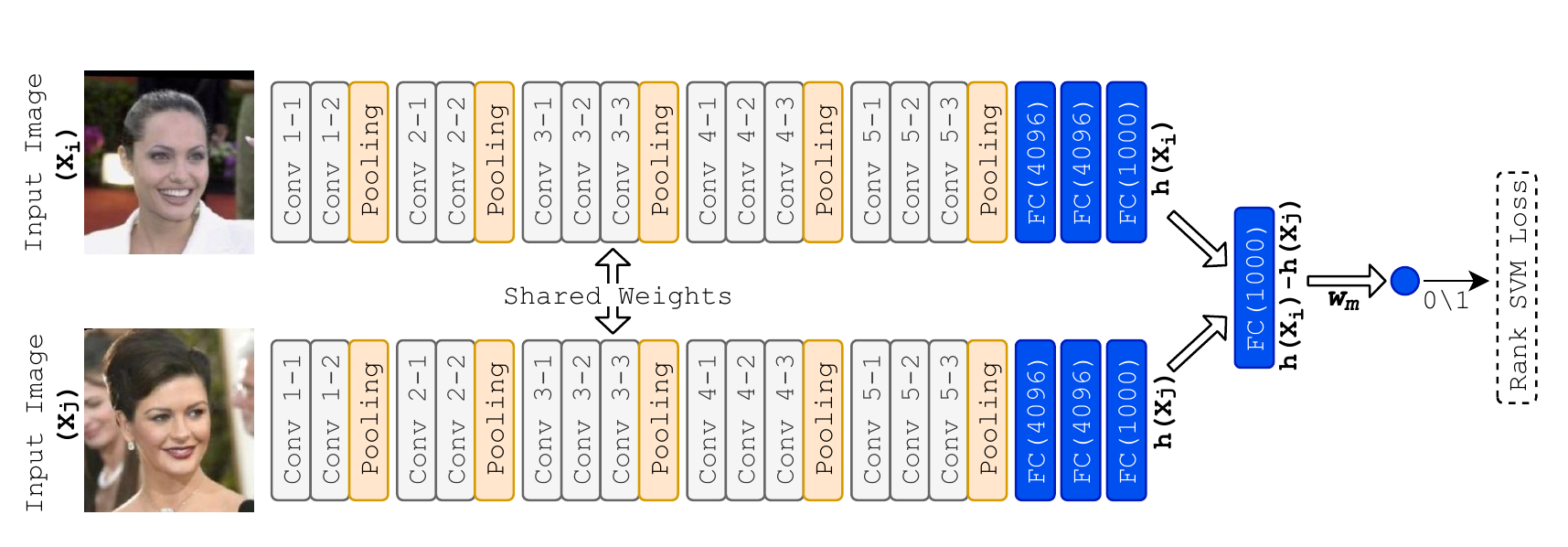}
    \caption{Overall Deep Rank SVM architecture. The network for a given attribute takes a pair of images ($\mathbf{x_i}$, $\mathbf{x_j}$) as input and outputs 1 if $\mathbf{x_i}$ shows the given attribute more strongly, compared to $\mathbf{x_j}$; 0 otherwise.}
    \label{fig:DeepRankFramework}
\end{figure*}

\section{Related Works}\label{sec:relatedWorks}

The relative attribute learning problem has attracted significant attention, with researchers approaching it first using traditional approaches and later using deep learning approaches.
Traditional and hand-crafted features are first used in
%for the relative attributes problem 
\cite{parikh2011relative,li2012relative,yu2014fine,xiao2015discovering,yu2015just}; however, more recently, deep convolutional neural networks are used to jointly learn the features and the ranking function in an end-to-end fashion \cite{souri2016deep,singh2016end,yang2016deep,zhang2018relative}. 
%We briefly review previous work below.

\subsection{Traditional Approaches}
Parikh and Grauman \cite{parikh2011relative}  first proposed relative attributes where they used the GIST descriptor \cite{oliva2001modeling} and color histogram features, together with a constrained optimization formulation similar to that of SVMs. %function is trained for every attribute based on the extracted features. 

Later, Li et al. \cite{li2012relative}, introduced non-linearity by using the relative 
forest algorithm to capture more accurate semantic relationships. More recently, Yu and Grauman \cite{yu2015just}, developed a Bayesian local learning strategy to infer when images are indistinguishable for a given attribute  in a probabilistic, local learning manner. 

\subsection{Deep Learning Approaches}
Hand-crafted feature representation may not be the best to capture the relevant visual features to describe relative attributes.
As with many other visual recognition problems, deep learning approaches significantly outperform approaches that are based on hand-crafted features followed by shallow  models.
%in relative attribute prediction due to the incorporation of deep representation learning. 

Souri et al. \cite{souri2016deep} introduced {RankNet}, which is  a convolutional neural network based on the architecture of VGG-16 \cite{simonyan2014very}.
%, adopted to learn deep features. 
A ranking layer is included to rank the strength of an attribute in the given pair of images based on the extracted features in an end-to-end fashion. 

Using a similar approach, Yang et al. \cite{yang2016deep} proposed a DRA model which consists of five convolutional neural layers and five fully connected layers followed by a relative loss function.

Singh and Lee \cite{singh2016end} trained a Siamese network based on AlexNet \cite{NIPS2012_4824}, with a pairwise ranking loss. The network consists of two branches, each branch consists of a localization module and a ranking module.
%I WOULD ADD A LINE HERE TOO
% I tried hocam, but I could not, they are just using a different technique or different loss function, like in Souri, he is using cross entropy, I can not say that cross entropy is bad. Actually, this is our message, to say rank SVM is better than cross entropy in this case.
%Singh and Lee is also using cross entropy + localization, it is making training/testing slower but it is not bad because I think if we have such thing + rank SVM instead of cross entropy, it can also further improve the accuracy.

In Zhuang et al. \cite{zhang2018relative}, cross-image representation is considered via deep attentive cross-image representation learning (DACRL) model: an end-to-end convolutional neural
network which takes a pair of images as input, and outputs a posterior probability that
indicates the relative strengths of a specific attribute, based on cross-image representation learning. 

Our work most resembles \cite{souri2016deep}, except for the loss function. As one of our contributions is embedding the SVM loss into the network, we compare our results to this system as the baseline; as well as newer work that achieved state-of-art results \cite{singh2016end} \cite{zhang2018relative}.

%%%%%%%%%%%%%%%%%%%%%%%%%%%%%%%%%%%%%%%
\section{Deep Rank SVM}
\label{sec:DRSVM}
We introduce Deep Rank SVM (DRSVM), 
a convolutional Siamese network trained with the rank SVM objective function. 
While rank SVM formulation was proposed before \cite{Thorsten2002}, this is the first time it is incorporated into a deep network architecture as a loss, to the best of our knowledge.

Following Parikh and Grauman's \cite{parikh2011relative} notation, training images consist of a set of ordered image-pairs $\mathbf{O_m} = {(\mathbf{x_i}, \mathbf{x_j})}$ and a set of un-ordered image-pairs $\mathbf{S_m} = {(\mathbf{x_i}, \mathbf{x_j})}$ for every attribute $m$ of a set of $M$ attributes. $(\mathbf{x_i}, \mathbf{x_j}) \in \mathbf{O_m}$ when the presence of attribute $m$ in $\mathbf{x_i}$ is stronger than the presence of attribute $m$ in $\mathbf{x_j}$ and $(\mathbf{x_i}, \mathbf{x_j}) \in \mathbf{S_m}$ when $\mathbf{x_i}$ and  $\mathbf{x_j}$ have similar presence strength of attribute $m$.

With these notations, we can formulate the  problem as learning the deep attribute representation $\mathbf{h(x)}$ 
of an image, for a specific attribute $m$ that satisfies the following constraints:
\begin{equation} \label{eq:rank}
\begin{array}{cc}
\mathbf{w^T_m}\mathbf{h(x_i)} > \mathbf{w^T_m}\mathbf{h(x_j)}; & \forall {(\mathbf{x_i}, \mathbf{x_j})} \in \mathbf{O_m}\\
\mathbf{w^T_m}\mathbf{h(x_i)} = \mathbf{w^T_m}\mathbf{h(x_j)}; & \forall {(\mathbf{x_i}, \mathbf{x_j})} \in \mathbf{S_m}
\end{array}
\end{equation}

In this work, we use the VGG-16 architecture \cite{simonyan2014very} as the base of a Siamese network to jointly learn the deep attribute representation $\mathbf{h(x)}$ and the weights $\mathbf{w_m}$ to rank 
the two input images for the given attribute $m$. 
The  network  is illustrated in Figure \ref{fig:DeepRankFramework}.
As seen in this figure, the output of the two branches of the network are  $1,000$ dimensional each. 
An additional layer is added to carry out the difference between the feature representations, $\mathbf{h(x_i)}$ and $\mathbf{h(x_j)}$; followed by an output node that computes the weighted differences between the two representations.
%We add a final layer to this network to learn to rank these deep representations of the two images.

%
For the objective function, we use the rank SVM formulation  proposed in \cite{parikh2011relative}; however unlike their use of the GIST features, we aimed to jointly learn the visual features and the rank SVM function, in a deep convolutional network.
%network parameters, while at the same time maximizing the distance between the ordered relative attributes. 
%
%The VGG-16 network is the 1st runner-up of the ILSVRC Competition in 2014 and is used here for comparison to the baseline system.
Details of the whole architecture are described in Section \ref{sec:ImpDetails}.

The input to the rank SVM function is the deep attribute representations $\mathbf{h(x_i)}$ and $\mathbf{h(x_j)}$, computed by the Siamese network for the image-pair, $\mathbf{x_i}$ and $\mathbf{x_j}$.
The rank SVM optimization function for relative attributes is defined  as in \cite{parikh2011relative}:
\begin{equation}
\label{eq:lossFunc}
\begin{aligned}
& \underset{}{\text{min}}
& & \dfrac{1}{2} \mathbf{w^T_m} \mathbf{w_m} + C_1 \textstyle{\sum} \xi_{ij}^2  + C_2 \textstyle{\sum} \gamma_{ij}^2 & \\
& \text{subject to}
& & \mathbf{w^T_m} (\mathbf{h(x_i)} - \mathbf{h(x_j)}) \geq 1 - \xi_{ij} & \forall(i, j) \in \mathbf{O_m} \\
& & & |\mathbf{w^T_m} (\mathbf{h(x_i)} - \mathbf{h(x_j)})| \leq \gamma_{ij} & \forall(i, j) \in \mathbf{S_m} \\
& & & \xi_{ij} \geq 0, \gamma_{ij} \geq 0 & 
\end{aligned}
\end{equation}

where $\mathbf{w_m}$ is the trainable weights of the ranking layer, 
%the first term is the network weight decay regularizer\ and the second and third terms are to enforce appropriate margin 
the first term maximizes the margin, while the second and third terms are there to enforce the soft margin of ordered/un-ordered image-pairs on the training images. 
We also used quadratic terms for the soft error as in \cite{parikh2011relative}.
$C_1$ and $C_2$ are the trade-off constants between maximizing
the margin and satisfying the pairwise relative constraints. We choose $C_1$ and $C_2$ to be equal  as done in \cite{parikh2011relative} and with the value of $0.1$.

We then obtain the corresponding unconstrained optimization problem by combining the constraints on the slack variables $\xi_{ij}$ and $\gamma_{ij}$, as:
%shown in Equation \ref{eq:unconstrained}:
%of Equation \ref{eq:unconstrained} is as follows:
\begin{equation}
\label{eq:unconstrained}
\begin{aligned}
\underset{\mathbf{w_m}}{\text{min}} ~~~\ignore{\quad} \dfrac{1}{2} \mathbf{w^T_m} \mathbf{w_m}  & + C_1\smashoperator{\sum_{(i,j) \in \mathbf{O_m}}}\max{(0, 1 - \mathbf{w^T_m} (\mathbf{h(x_i)} - \mathbf{h(x_j)}))^2} \\
& + C_2\smashoperator{\sum_{(i,j) \in \mathbf{S_m}}} (\mathbf{w^T_m} (\mathbf{h(x_i)} - \mathbf{h(x_j)})^2
\end{aligned}
\end{equation}

%Additionally, we would like to emphasise about the usage of bias term in the objective function of rank SVM. Networks with bias terms can easily learn any constant function, regardless of the input of the network. It follows that bias terms should not be used in neural networks with Deep Rank SVM as the network can learn the constant function mapping directly to the relative ordering.

%Finally, the whole network, Siamese network with rank SVM function, is jointly trained and optimized as shown in Figure \ref{fig:DeepRankFramework}. 

As suggested in \cite{pmlr-v80-ruff18a}, the objective function is calculated with no bias term to avoid learning the constant function mapping directly to the relative ordering.

\section{Experimental Evaluation}
\label{sec:ExpRes}

%\subsection{Comparison to State-of-Art}
%\label{baseline}
We evaluate the effectiveness of our approach on the publicly available  datasets for relative attributes, described in Section \ref{sec:datasets}.
Our results are compared to the results of several systems that report accuracy as performance measurement,
namely Rank SVM \cite{parikh2011relative}, FG-LP \cite{yu2014fine}, spatial Extent \cite{xiao2015discovering}, DeepSTN \cite{singh2016end}, DRA \cite{yang2016deep}, and DACRL \cite{zhang2018relative}.

We consider the {RankNet} system proposed in \cite{souri2016deep} as our baseline. We used the same network (VGG-16 pre-trained on ILSVRC 2014) and the same data augmentation techniques, but the proposed rank SVM loss function was used instead of the cross-entropy loss used in {RankNet}. In this way we aimed to evaluate the effectiveness of using the SVM formulation with our deep learning framework. 

In Section \ref{sec:ImpDetails}, the network and implementation details are explained. 
%Section \ref{baseline} discusses the most similar work to ours, used as the baseline. 
In Section \ref{sec:Res}, the performance of
our proposed method is shown along with a comparison with the baseline and other state-of-the-art systems.

\subsection{Datasets}
\label{sec:datasets}
Our proposed approach is evaluated on four different datasets from distinctive areas for comprehensive evaluation.
\begin{enumerate}
    \item \textbf{LFW-10 Dataset \cite{sandeep2014relative}}: The dataset is a subset of the Labels Faces in the Wild (LFW) dataset \cite{LFWTech}. It consists of $2,000$ images with $10$ different face attributes (refer to Table \ref{tbl:LFWRes}). For each attribute, a random subset of 500 pairs of images have been annotated for training and testing sets.

    \item \textbf{Public Figure Face Dataset \cite{parikh2011relative}}: PubFig dataset consists of $772$ images from $8$ different identities with $11$ semantic attributes (refer to Table \ref{tbl:PubFigRes}). The ordering of the samples are annotated at the category level; in other words, all images with the same identity are ranked higher, equal, or lower than all images belonging to another identity with respect to a specific attribute. For instance, a person is said to have longer hair than another person, even if this may not be true in all of their photographs. This short-cut in annotation will result in label inconsistencies.

    \item \textbf{UTZap50K-2 Dataset \cite{yu2014fine}}: Large shoe dataset consists of $50,025$ images collected from Zappos.com. It consists of 4 shoe attributes: open, pointy, sporty, and comfort (refer to Table \ref{tbl:ZapposRes}). After pruning out pairs with low confidence or agreement, the human-annotated examples consist of approximately $1,500$-$1,800$ training image-pairs for each attribute and in total $4,334$ image-pairs for testing. 
    
    \item \textbf{Zappos50K-lexi Dataset \cite{yu2017semantic}}: It is based on UTZap50K dataset \cite{yu2014fine} with 10 additional  fine-grained relative attributes: comfort, casual, simple, sporty, colorful, durable, supportive, bold, sleek, and open (refer to Table \ref{tbl:ZapposLexiRes}). The dataset consists of approximately $1,300$-$2,100$ image-pairs for each attribute.
\end{enumerate}
\noindent
In all of our experiments, we have used the provided training/testing split by the original publishers of the datasets.

\begin{table*}[]
\centering
\begin{tabular}{|p{2.7cm}|p{0.85cm} p{0.85cm} p{0.85cm} p{0.85cm} p{0.85cm} p{0.85cm} p{0.85cm} p{0.85cm} p{0.85cm} p{0.85cm} |p{1.2cm}|}
\hline
Method & \begin{tabular}[c]{@{}c@{}}Bald\\ Head\end{tabular} & \begin{tabular}[c]{@{}c@{}}Dark\\ Hair\end{tabular} & \begin{tabular}[c]{@{}c@{}}Eyes\\ Open\end{tabular} & \begin{tabular}[c]{@{}c@{}}Good\\ \small{Looking}\end{tabular} & \begin{tabular}[c]{@{}c@{}}Mascu.\\ \small{Looking}\end{tabular} & \begin{tabular}[c]{@{}c@{}}Mouth\\ Open\end{tabular} & Smile & Teeth & \begin{tabular}[c]{@{}c@{}}Fore-\\ head\end{tabular} & Young &{\hspace{0.05cm} Mean} \\ \hline
FG-LP \cite{yu2014fine}& 67.90 & 73.60 & 49.60 & 64.70 & 70.10 & 53.40 & 59.70 & 53.50 & 65.60 & 66.20 & 62.43\% \\ \hline
Spatial Extent \cite{xiao2015discovering} & 83.21 & 88.13 & 82.71 & \textbf{72.76} & 93.68 & 88.26 & 86.16 & 86.46 & 90.23 & 75.05 & 84.67\% \\ \hline
RankNet \cite{souri2016deep}& 81.14 & 88.92 & 74.44 & 70.28 & 98.08 & 85.46 & 82.49 & 82.77 & 81.90 & 76.33 & 82.18\% \\ \hline
DeepSTN  \cite{singh2016end}& 83.94 & 92.58 & \textbf{90.23} & 71.21 & 96.55 & 91.28 & 84.75 & 89.85 & 87.89 & 80.81 & 86.91\% \\ \hline
\begin{tabular}[c]{@{}l@{}}DACRL \cite{zhang2018relative}\\ (without Attention)\end{tabular} & 83.21& 91.99& 87.97& 69.97& 97.70& 89.93& 85.03& 88.00& 89.45& 74.84& 85.81\% \\ \hline 
DACRL  \cite{zhang2018relative}& 85.04 & 92.58 & \textbf{90.23} & 70.28 & \textbf{98.28} & 91.28 & 85.03 & 89.23 & 90.63 & 76.55 & 86.91\% \\ \hline
DRSVM   (proposed)       &\textbf{90.75}& \textbf{92.67} & 86.54 & 71.21 & 95.05 & \textbf{92.67} & \textbf{88.64} & \textbf{91.80} & \textbf{90.84} & \textbf{81.02} & \textbf{88.12\%} \\ \hline 
\end{tabular}
\caption{State-of-the-art accuracies on LFW-10 dataset compared with the results
obtained in this work. Bold figures indicate the best results.}
\label{tbl:LFWRes}
\end{table*}

%%%%%%%%%%%%%%%%%%%%%%%%%%%%%%%%%%%%%%%%% PubFig %%%%%%%%%%%%%%%%%%%%%%%%%%%%%%%%%%%%%%%%%%%%%%%%%
\begin{table*}[]
\centering
\begin{tabular}{|p{2.3cm}|p{0.8cm} p{0.8cm} p{0.8cm} p{0.8cm} p{0.8cm} p{0.8cm} p{0.8cm} p{0.8cm} p{0.8cm} p{0.8cm} p{0.8cm} |p{1.2cm}|}
\hline
Method & Male & White & Young & Smile & Chubby & \begin{tabular}[c]{@{}c@{}}Fore-\\head\end{tabular} & \begin{tabular}[c]{@{}c@{}}Bushy \\
\small{Eyebrow} \end{tabular}& \begin{tabular}[c]{@{}c@{}}\small{Narrow} \\ Eyes \end{tabular} & \begin{tabular}[c]{@{}c@{}}\small{Pointy}\\Nose\end{tabular} & \begin{tabular}[c]{@{}c@{}}Big\\Lips\end{tabular} & \begin{tabular}[c]{@{}c@{}}Round\\Face\end{tabular} & {\hspace{0.09cm}Mean} \\ \hline
FG-LP \cite{yu2014fine}       & 91.77& 87.43& 91.87& 87.00& 87.37& 94.00& 89.83& 91.40& 89.07& 90.43& 86.70& 89.72\%\\ \hline
RankNet \cite{souri2016deep}     & 95.50& 94.60& 94.33& 95.36& 92.32& 97.28& 94.53& 93.19& 94.24& 93.62& 94.76& 94.42\%\\ \hline
DRA  \cite{yang2016deep}        & 90.82& 87.12& 91.49& 92.68& 89.30& 94.39& 90.19& 90.60& 91.03& 90.35& 91.99& 90.91\%\\ \hline
\begin{tabular}[c]{@{}l@{}}DACRL \cite{zhang2018relative}\\ (without Attention)\end{tabular} & 97.70& 97.82 &97.10& 97.03& 97.05& 98.30& 97.36 &97.99& 97.26& 94.36& 98.04& 97.27\%\\ \hline
DACRL \cite{zhang2018relative}      & 96.49& 97.80& \textbf{97.96}& \textbf{97.42}& \textbf{97.22}& 98.05& 97.48& 96.91& 97.74& 96.83& 96.27& 97.29\%\\ \hline
DRSVM (proposed)        & \textbf{97.74} & \textbf{98.61} & 96.32 & 96.14 & 94.47 & \textbf{98.75}&\textbf{98.68} & \textbf{97.28} & \textbf{99.31} & \textbf{98.24} & \textbf{96.89} & \textbf{97.49}\%\\ \hline
\end{tabular}
\caption{Comparison of the state-of-the-art accuracies on the PubFig dataset.}
\label{tbl:PubFigRes}
\end{table*}

%%%%%%%%%%%%%%% ZapposLexi %%%%%%%%%%%%%%%%%%%%
\begin{table*}[]
\centering
\begin{tabular}{|p{2.3cm}|p{0.8cm} p{0.8cm} p{0.8cm} p{0.8cm} p{1cm} p{0.8cm} p{1cm} p{0.8cm} p{0.8cm} p{0.8cm} |p{1.2cm}|}
\hline
Method & Comfort & Casual & Simple & Sporty & Colorful & Durable & Supportive & Bold & Sleek & Open & {\hspace{0.09cm}Mean} \\ \hline

RankNet \cite{souri2016deep} & 90.48 & 90.43 & 90.40 & 93.31 & 95.43 & 90.47 & 91.98 & 91.53 & 86.31 & 82.53 & 90.29\%\\ \hline

\begin{tabular}[c]{@{}l@{}}DACRL \cite{zhang2018relative}\\ (without Attention)\end{tabular} & \textbf{91.88} & 94.44 & 89.93 & 93.01 & \textbf{97.33} & 92.65 & 92.65 & 91.12 & 89.24 & 87.90 & 92.02\%\\ \hline

DACRL \cite{zhang2018relative}  & \textbf{91.88} & 91.36 & 90.16 & 94.22 & 95.81 & 92.33 & 92.65 & \textbf{92.56} & \textbf{90.71} & 88.98 & 92.07\%\\ \hline

DRSVM (proposed) & 91.59 & \textbf{95.37} & \textbf{90.91} & \textbf{96.57} & 95.95 & \textbf{93.31} & \textbf{94.98} & 91.47 & 89.30 & \textbf{89.99}& \textbf{92.94\%}\\ \hline

%91.59+95.37+90.91+96.57+95.95+93.31+94.98+ 91.47+89.30+89.99 = 929.44/10 = 92.94
\end{tabular}
\caption{Comparison of the state-of-the-art accuracies on the UTZap50K-lexi dataset.}
\label{tbl:ZapposLexiRes}
\end{table*}

%%%%%%%%%%%%%%% Zappos50K-2 %%%%%%%%%%%%%%%%%%%%
\begin{table*}[ht]
\centering
\begin{tabular}{|l|c c c c|c|}
\hline
Method      & Open&Pointy&Sporty&Comfort & Mean \\ \hline 
RankSVM \cite{parikh2011relative}    & 60.18 & 59.56 & 62.70 & 64.04 & 61.62\% \\ \hline
FG-LP   \cite{yu2014fine}    & 74.91 & 63.74 & 64.54 & 62.51 & 66.43\% \\ \hline
RankNet \cite{souri2016deep}    & 73.45 & 68.20 & 73.07 & 70.31 & 71.26\% \\ \hline
\begin{tabular}[c]{@{}l@{}}DACRL \cite{zhang2018relative}\\ (without Attention)\end{tabular} & 75.45& 69.80 &73.78 &68.54& 71.89\% \\ \hline
DACRL   \cite{zhang2018relative}    & \textbf{75.66} & 70.65 & \textbf{73.87} & 69.56 & \textbf{72.44}\% \\ \hline 
DRSVM (proposed)    & 74.09 & \textbf{70.90} & 72.95 & \textbf{71.20} & 72.29\% \\ \hline
\end{tabular}
\caption{Comparison of the state-of-the-art accuracies on the UTZap50K-2 dataset.}
\label{tbl:ZapposRes}
\end{table*}

\subsection{Implementation Details}
\label{sec:ImpDetails}
We chose the VGG-16 model to be our base architecture to have a fair comparison with our baseline \cite{souri2016deep}, which uses the same architecture. 

The input to the model is two $224\times224$ RGB images similar to \cite{souri2016deep}.
The output of the two branches of Siamese network $\mathbf{h(x_i)}$ and $\mathbf{h(x_j)}$ are $1,000$ dimensional each, as illustrated in Figure \ref{fig:DeepRankFramework}. 
The  weight initialization for the output node is sampled from a zero-mean Gaussian distribution with a standard deviation of $0.01$. 

To implement and fine-tune our architecture, we have used the pre-trained VGG-16 provided by Keras framework. %The pre-trained weights are used as initialization and fine-tuned to predict the relative ordering. 
The last added weights $\mathbf{w_m}$ of the ranking layer is initialized using the Xavier method without bias term. For training, stochastic gradient descent with RMSProp optimizer is used with a mini-batch
size of $48$ image-pairs. A unified learning rate is set to $10^{-5}$ for all of the layers. The training images are shuffled after every epoch. 

A separate network for each attribute is trained. For LFW-10, UTZap50K-2, and UTZap50K-lexi datasets, we trained our model for 500, 200, and 200 epochs for each attribute, respectively. For PubFig dataset, the relative attributes are annotated at the category-level manner; hence one epoch contains large number of image-pairs, that is the combination of all of the images in the dataset. Therefore, we trained the model for $10,000$ iterations where in every iteration a random selection of 48 image-pairs are chosen from the dataset, and ground-truth labels are assigned based on their categories.

Advanced data augmentation techniques have proven to improve performance in many studies specially for deep learning. However, to resemble our baseline and show the effectiveness of incorporating Rank SVM loss function to the deep learning, only simple on-the-fly data augmentation techniques are applied during training, namely rotation [-15, 15], horizontal flipping, and random cropping.

%%%%%%%%%%%%%%%%%%%%%%%%%%%%%%%%%%%%%%%%%%%%%%%%%%%%%%%%%%
\subsection{Results}
\label{sec:Res}
We compare the performance of the proposed Deep Rank SVM model (DRSVM) with our baseline \cite{souri2016deep} and state-of-the-art, on four different datasets, in Tables \ref{tbl:LFWRes}-\ref{tbl:ZapposRes}.
The reported performance figures are accuracies over correctly ordered pairs (excluding similar pairs), in line with the literature.

Table \ref{tbl:LFWRes} shows the results on the LFW-10 dataset where we outperform our baseline \cite{souri2016deep} by about $6\%$ on average. 
We can attribute this to the use of the rank SVM loss as the loss function, as this is the main difference between our  model and the baseline.
Our results surpass the average accuracy by $1.2\%$ points over the state-of-the-art \cite{zhang2018relative}, with best performance on $7$ of the  $10$ attributes.

Table \ref{tbl:PubFigRes} shows the results on the PubFig dataset where our system improves over the baseline \cite{souri2016deep} by $3\%$ points and obtains the best results on $8$ out of $11$ attributes compared to state-of-art \cite{zhang2018relative}. The gain on this dataset is marginal ($0.2\%$) which may be due to the category base annotation that may result in annotation inconsistencies, as discussed in Section \ref{sec:datasets}. 

Table \ref{tbl:ZapposLexiRes} shows the results on the UTZap50K-lexi dataset where we outperform the baseline by $2.65\%$. Furthermore, we improve the average accuracy by $0.87\%$ over the state-of-the-art \cite{zhang2018relative} and obtain the best results in 6 out of the 10 attributes. 

Finally, Table \ref{tbl:ZapposRes} shows the results on the UTZap50K-2 dataset where we outperform the baseline by $1\%$,
but slightly underperform the state-of-art \cite{zhang2018relative} by $0.15\%$ 
($72.29\%$ vs $72.44\%$), while obtaining best results in 2 out of 4 attributes.

%%%%%%%%%%%%%%%%%%%%%%%%%%%%%%%%%%%%%%%%%%%%%%%%%%%%%%%%%%
\subsection{Discussion}
The reported results in Tables \ref{tbl:LFWRes}-\ref{tbl:ZapposLexiRes} show that we outperformed our baseline \cite{souri2016deep} on the four employed datasets, LFW-10, PubFig, UTZap50K-lexi, and UTZap50K-2 by $6\%$, $3\%$, $2.65\%$, and $1\%$ points respectively. This shows the effectiveness of using the rank SVM loss with the deep learning approach, for the relative attribute learning problem.

Furthermore, we surpassed the state-of-art on the  LFW-10, PubFig, and UTZap50K-lexi datasets by $1.2\%$, $0.2\%$, and $0.87\%$ points and obtained slightly lower results on the UTZap50K-2 dataset ($72.44\%$ versus $72.29\%$). 

To show the effectiveness of incorporating the rank SVM objective function into our deep learning framework, we employed the same  architecture used in our baseline \cite{souri2016deep} and in DACRL \cite{zhang2018relative}, namely VGG-16.
We expect that the performance of our model will be even higher with a more advanced network (e.g. Inception-ResNet [28] or NasNetLarge [29]) and using heavy data augmentation.

%We believe that this slightly worse performance on  the UTZap50K-2 dataset  can be explained as follows. To show the effectiveness using the rank SVM objective function in our deep learning model, we employed the same deep architecture used in our baseline \cite{souri2016deep} and in DACRL \cite{zhang2018relative}, namely VGG-16. However, DACRL has the advantage of incorporating the channel-wise attention of concatenated single-image feature maps to their architecture. Therefore, for a fair comparison, the performance of DACRL without incorporating channel-wise attention is also included in Tables \ref{tbl:LFWRes}, \ref{tbl:PubFigRes}, \ref{tbl:ZapposRes}, and \ref{tbl:ZapposLexiRes}. The results show that our system outperforms this version of DACRL by $2.31\%$, $0.22\%$, $0.4\%$, and $0.92\%$ on LFW-10, PubFig, UTZap50K-2, and UTZap50K-lexi datasets, respectively. 
%Furthermore, we expect the performance of our system to be  better with a more advanced network as the base (instead of the VGG-16) and heavy data augmentation.

\begin{centering}
\begin{figure*}
\centering
     \caption*{(a) LFW10 (Bald Head attribute)}
     \label{fig:LFW_BaldHead}
    \subfloat[-0.9716]{\includegraphics[width=0.115\linewidth]{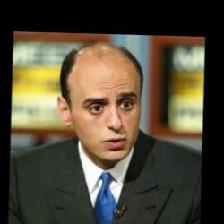}}   \hspace{0.01cm}
    \subfloat[-0.3155]{\includegraphics[width=0.115\linewidth]{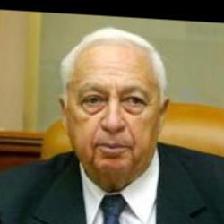}}  \hspace{0.01cm}
    \subfloat[-0.0568]{\includegraphics[width=0.115\linewidth]{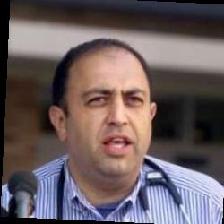}} \hspace{0.01cm}
    \subfloat[0.0339]{\includegraphics[width=0.115\linewidth]{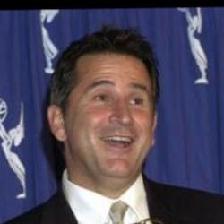}}  \hspace{0.01cm}
    \subfloat[0.3480]{\includegraphics[width=0.115\linewidth]{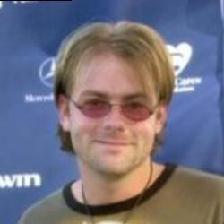}}  \hspace{0.01cm}
    \subfloat[0.4025]{\includegraphics[width=0.115\linewidth]{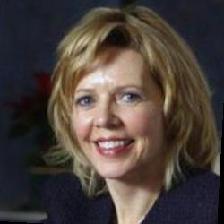}}  \hspace{0.01cm}
    \subfloat[0.8651]{\includegraphics[width=0.115\linewidth]{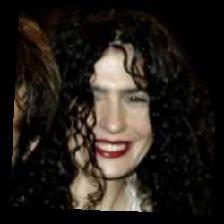}}  \hspace{0.01cm}
    \subfloat[1.0]{\includegraphics[width=0.115\linewidth]{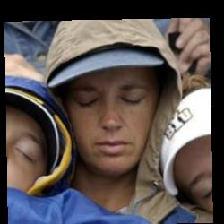}}

    \hfill
    \caption*{(b) PubFig (Smile attribute)}
    \label{fig:PubFig_Smile}
    \subfloat[-1.0]{\includegraphics[width=0.115\linewidth]{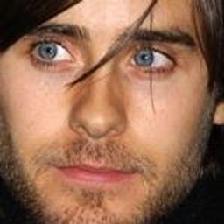}}  \hspace{0.01cm}
    \subfloat[-0.4188]{\includegraphics[width=0.115\linewidth]{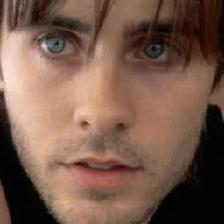}}  \hspace{0.01cm}
    \subfloat[-0.2701]{\includegraphics[width=0.115\linewidth]{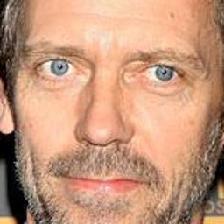}}  \hspace{0.01cm}
    \subfloat[-0.0933]{\includegraphics[width=0.115\linewidth]{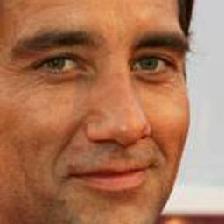}}  \hspace{0.01cm}
    \subfloat[0.3018]{\includegraphics[width=0.115\linewidth]{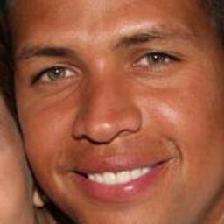}}  \hspace{0.01cm}
    \subfloat[0.4799]{\includegraphics[width=0.115\linewidth]{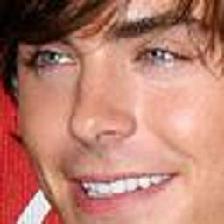}}  \hspace{0.01cm}
    \subfloat[0.6231]{\includegraphics[width=0.115\linewidth]{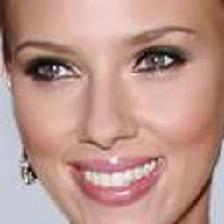}}  \hspace{0.01cm}
    \subfloat[0.8195]{\includegraphics[width=0.115\linewidth]{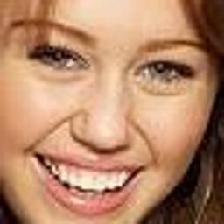}}  \hspace{0.01cm}

    \hfill
    \caption*{(c) UTZap50K-2 (Open attribute)}
    \label{fig:Zappos_Open}
    \subfloat[-1.0]{\includegraphics[width=0.115\linewidth]{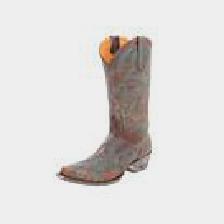}}  \hspace{0.01cm}
    \subfloat[-0.6688]{\includegraphics[width=0.115\linewidth]{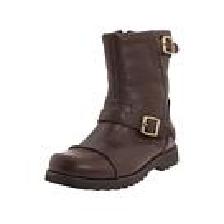}}  \hspace{0.01cm}
    \subfloat[-0.3944]{\includegraphics[width=0.115\linewidth]{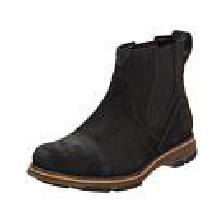}}  \hspace{0.01cm}
    \subfloat[-0.1453]{\includegraphics[width=0.115\linewidth]{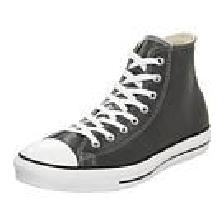}}  \hspace{0.01cm}
    \subfloat[0.1414]{\includegraphics[width=0.115\linewidth]{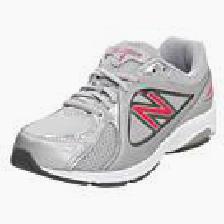}}  \hspace{0.01cm}
    \subfloat[0.3243]{\includegraphics[width=0.115\linewidth]{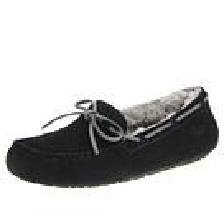}}  \hspace{0.01cm}
    \subfloat[0.5513]{\includegraphics[width=0.115\linewidth]{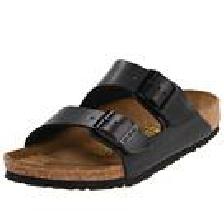}}  \hspace{0.01cm}
    \subfloat[1.0]{\includegraphics[width=0.115\linewidth]{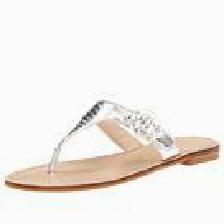}}
    
\caption{Sample images ordered according to their output prediction of their associated attribute.}
\label{fig:OrderedSamples}
\end{figure*}
\end{centering}

Figure \ref{fig:OrderedSamples} shows some images along with their output prediction values of the respective attribute. Although, our network is trained given only image-pairs, we can see that the network has learned  a global ranking of a given attribute. 

\begin{centering}
\begin{figure*}
\centering
\caption*{}
\begin{minipage}{.12\textwidth} \begin{tabular}[c]{@{}c@{}}  Original\\  \vspace{-2.4cm} Images\end{tabular} \end{minipage}\hspace{0.05cm}
\subfloat[Bald Head]{\includegraphics[width=0.13\linewidth]{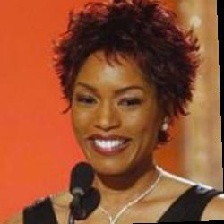}}\hspace{0.05cm}
\subfloat[Teeth]{\includegraphics[width=0.13\linewidth]{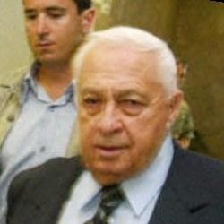}}\hspace{0.05cm}
\subfloat[Bushy Eyebrows]{\includegraphics[width=0.13\linewidth]{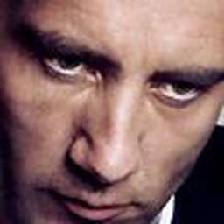}}\hspace{0.05cm}
\subfloat[Pointy Nose]{\includegraphics[width=0.13\linewidth]{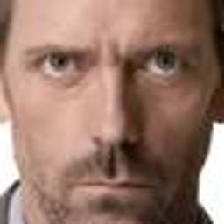}}\hspace{0.05cm}
\subfloat[Open]{\includegraphics[width=0.13\linewidth]{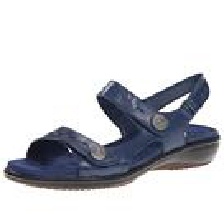}}\hspace{0.05cm}
\subfloat[Pointy]{\includegraphics[width=0.13\linewidth]{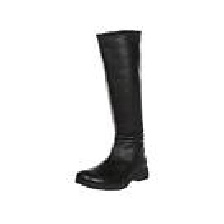}}

\vspace{-0.5cm}
\hfill %%%%%%%%%%%%%%%%%%%%%%%%%%%%%%%%%%%%

\begin{minipage}{.12\textwidth} Heat Maps\vspace{-2.4cm} \end{minipage}\hspace{0.05cm}
\subfloat{\includegraphics[width=0.13\linewidth]{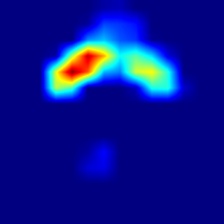}}\hspace{0.05cm}
\subfloat{\includegraphics[width=0.13\linewidth]{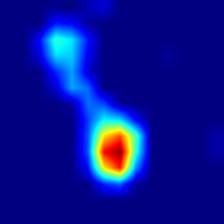}}\hspace{0.05cm}
\subfloat{\includegraphics[width=0.13\linewidth]{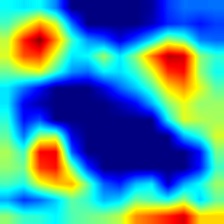}}\hspace{0.05cm}
\subfloat{\includegraphics[width=0.13\linewidth]{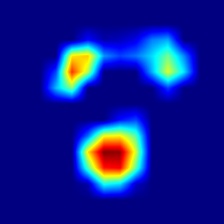}}\hspace{0.05cm}
\subfloat{\includegraphics[width=0.13\linewidth]{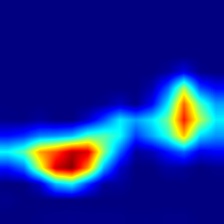}}\hspace{0.05cm}
\subfloat{\includegraphics[width=0.13\linewidth]{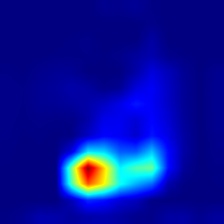}}

\vspace{-0.5cm}
\hfill %%%%%%%%%%%%%%%%%%%%%%%%%%%%%%%%%%%%

\begin{minipage}{.12\textwidth} \begin{tabular}[c]{@{}c@{}}  Superimposed\\  \vspace{-2.5cm} Images\end{tabular} \end{minipage}\hspace{0.05cm}
\subfloat{\includegraphics[width=0.13\linewidth]{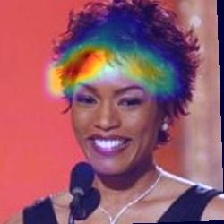}}\hspace{0.05cm}
\subfloat{\includegraphics[width=0.13\linewidth]{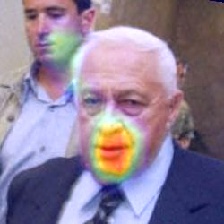}}\hspace{0.05cm}
\subfloat{\includegraphics[width=0.13\linewidth]{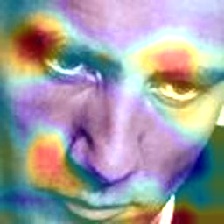}}\hspace{0.05cm}
\subfloat{\includegraphics[width=0.13\linewidth]{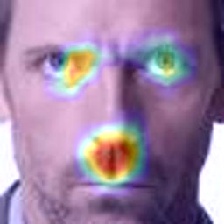}}\hspace{0.05cm}
\subfloat{\includegraphics[width=0.13\linewidth]{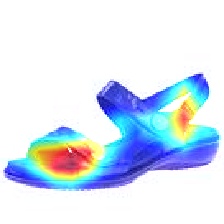}}\hspace{0.05cm}
\subfloat{\includegraphics[width=0.13\linewidth]{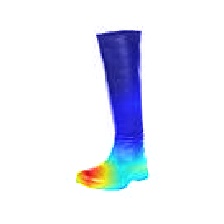}}\hspace{0.05cm}
 
\caption{Class Activation Maps showing the pixels with most contribution to the ranking prediction in (a) Bald Head and Teeth from LFW-10 dataset, (b) Bushy Eyebrows and Pointy Nose from PubFig dataset, and (c) Open and Pointy from UTZap50K-2 dataset.}
\label{fig:CAM}
\end{figure*}
\end{centering}

The trained network is able to localize on the informative regions of the image related to a given attribute, without explicitly being taught to do so  during training. 
We calculate the derivative of the output with respect to a given input and visualize the results of the last convolutional layer as shown in Figure \ref{fig:CAM}.
The heat maps visualize the pixels in the images with the most contribution to the ranking prediction of the network.

\section{Conclusions and Future Work}
\label{sec:conc}
In this paper, we proposed the deep rank SVM (DRSVM) network for relative attribute learning, to jointly learn the features and the ranking function in an end-to-end fashion. Our model is evaluated on four benchmarks, LFW-10, PubFig, UTZap50K-lexi and UTZap50K-2 and achieved state-of-the-art performance on LFW-10, PubFig, and UTZap50K-lexi datasets. 
These results shows the benefit of incorporating and jointly training the network with Rank SVM loss function for relative attributes.   

%% S: Is it fine to add references in conclusion? Himm * ok but not often done - or not too much
Although the results show the ability of the network to localize on the informative regions in the image, adding a localization module similar to the one used in \cite{zhang2018relative} can contribute to the performance,  especially in the existence of some annotation inconsistencies, as in the case of PubFig dataset.

We believe that the performance can be further improved with a better network (e.g. Inception-ResNet \cite{szegedy2017inception} or NasNetLarge \cite{zoph2018learning}) than the one used in this work (VGG-16), as well as using heavy data augmentation.
We will add results obtained with a more powerful network in the final version of the manuscript.
%we would like to investigate the effectiveness of attention mechanism in our model

Source code of the proposed method is provided in supplementary materials, and will be made public upon acceptance.

\section*{Acknowledgment}

This work as supported by a  grant from The Scientific and Technological Research Council of Turkey (TÜBİTAK) under project number 119E429. 

%Sara Atito Ali Ahmed was supported partially by a Sabanc{\i} University fellowship and during this work. 



%\bibliographystyle{IEEEtran}

\bibliographystyle{./IEEEtran}
\bibliography{./ms}

\end{document}